\documentclass[lettersize,journal]{IEEEtran}
\usepackage{amsmath,amsfonts}
\usepackage{algorithmic}
\usepackage{algorithm}
\usepackage{array}
\usepackage[caption=false,font=normalsize,labelfont=sf,textfont=sf]{subfig}
\usepackage{textcomp}
\usepackage{stfloats}
\usepackage{url}
\usepackage{verbatim}
\usepackage{graphicx}
\usepackage{cite}
\usepackage{amsmath,amsfonts, physics}
\usepackage{algorithmic}
\usepackage{array}
\usepackage[caption=false,font=normalsize,labelfont=sf,textfont=sf]{subfig}
\usepackage{textcomp}
\usepackage{stfloats}
\usepackage{url}
\usepackage{verbatim}
\usepackage{graphicx}
\usepackage{svg}
\usepackage{hyperref}
\usepackage{enumitem}
\usepackage{multirow}
\usepackage{amssymb}
\usepackage{float}
\usepackage{enumitem}
\newlist{steps}{enumerate}{1}
\setlist[steps, 1]{label = Step \arabic*:}
\begin{document}

\title{Towards Deep Learning Guided Autonomous Eye Surgery using Microscope and iOCT Images}

\author{Ji Woong Kim$^{1*}$, Shuwen Wei$^{2*}$, Peiyao Zhang$^{1*}$, \\ Peter Gehlbach$^{3}$, Jin U. Kang$^{2}$, Iulian Iordachita$^{1}$, Marin Kobilarov$^{1}$
    
\thanks{$^{1}$ J. W. Kim, P. Zhang, I. Iordachita, and M. Kobilarov are with the Mechanical Engineering Dept at the Johns Hopkins University, Baltimore, MD 21218 USA}
\thanks{$^{2}$ S. Wei and J. U. Kang is with the Electrical and Computer Engineering Dept at the Johns Hopkins University, Baltimore, MD 21218 USA}
\thanks{$^{3}$ P. Gehlbach is with the the Johns Hopkins Wilmer Eye Institute, Baltimore, MD 21287 USA}
\thanks{$^{*}$ denotes equal contribution}
}

\markboth{Journal of \LaTeX\ Class Files,~Vol.~14, No.~8, July~2023}%
{Shell \MakeLowercase{\textit{et al.}}: A Sample Article Using IEEEtran.cls for IEEE Journals}


\maketitle

\begin{abstract}

Recent advancements in retinal surgery have paved the way for a modern operating room equipped with a surgical robot, a microscope, and intraoperative optical coherence tomography (iOCT) – a depth sensor widely used in retinal surgery. Integrating these tools raises the fundamental question of how to effectively combine them to enable surgical autonomy. In this work, we tackle this question by developing a unified framework that facilitates real-time autonomous surgical workflows leveraging these devices. The system features: (1) a novel imaging system that integrates the microscope and iOCT in real-time by dynamically tracking the surgical instrument via a small iOCT scanning region, providing real-time depth feedback; (2) implementation of convolutional neural networks (CNN) that automatically detect and segment task-relevant information for surgical autonomy; (3) intuitive selection of goal waypoints within both the microscope and iOCT views through simple mouse-click interactions; and (4) integration of model predictive control (MPC) for trajectory generation, ensuring patient safety by implementing safety-related kinematic constraints. The system's utility is demonstrated by automating subretinal injection (SI), a challenging procedure with high accuracy and depth perception requirements. We validate our system by conducting 30 successful SI trials on pig eyes, achieving mean needle insertion accuracy of $26 \pm 12 \mu m$ to various subretinal goals and mean duration of  $55 \pm 10.8$ seconds. Preliminary comparisons to a human operator performing SI in robot-assisted mode highlight the enhanced safety of our system. \href{https://sites.google.com/view/eyesurgerymicroscopeoct/home
}{\color{blue}{Project website is here.}}

\begin{IEEEkeywords}
Vision-Based Navigation; Computer Vision for Medical Robotics; Medical Robots and Systems
\end{IEEEkeywords}
\end{abstract}

\section{Introduction}

\begin{figure}
\centering
\includegraphics[width = \columnwidth]{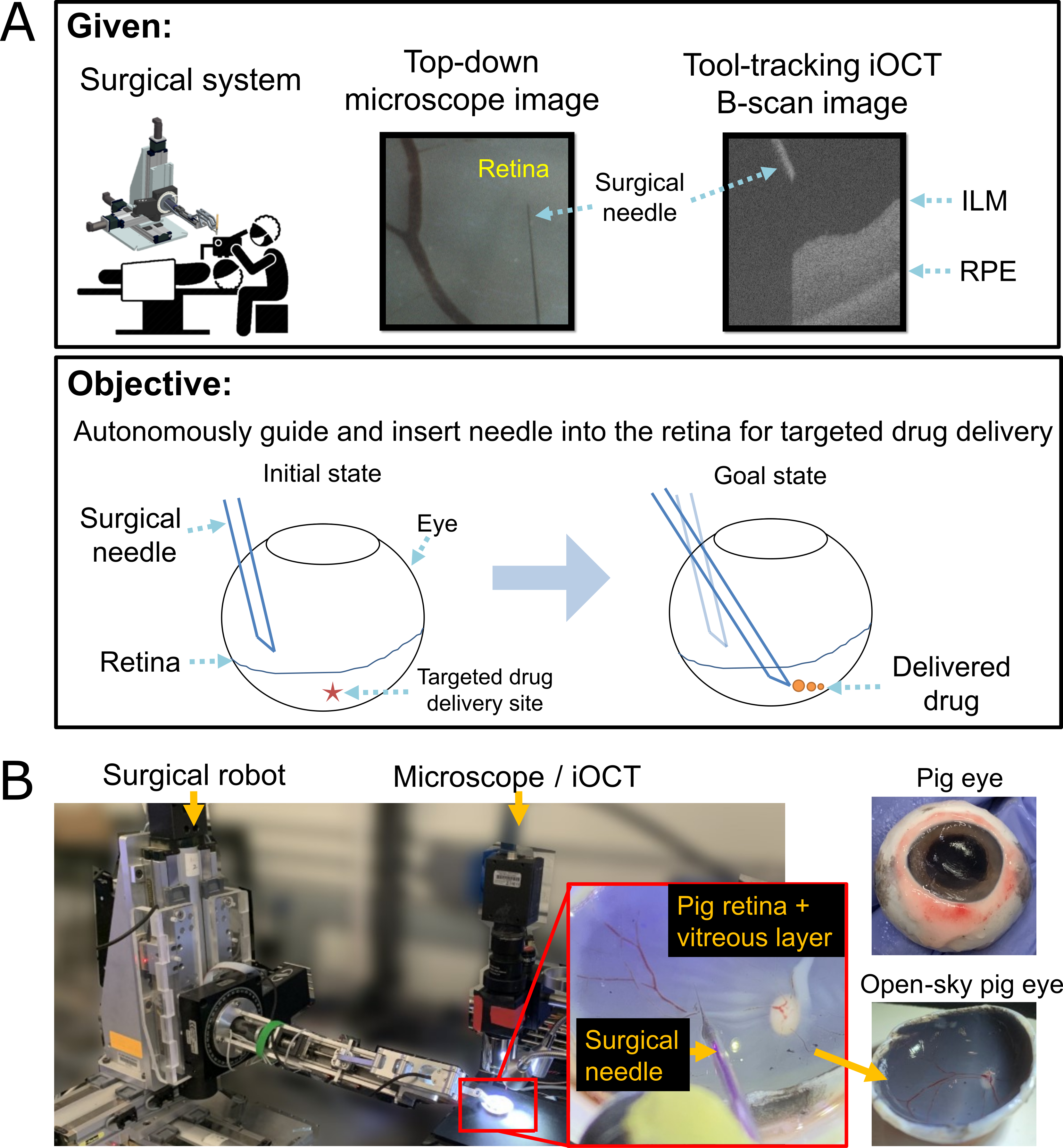}
\caption{\small (A) Problem statement (B) Experimental setup}
\label{fig:intro_fig}
\end{figure}

Subretinal injection (SI) is a surgical procedure that involves inserting a micro-needle between the retina's photoreceptor layer and the underlying retinal pigment epithelium (RPE) layer for targeted drug delivery (Fig. \ref{fig:intro_fig}). Unlike the standard intravitreal delivery method, which administer drugs above the retina and may not reach the desired subretinal space, SI offers greater effectiveness by delivering the drug in direct contact with the targeted subretinal tissue \cite{first_human_subretinal_injection}.  However, SI presents several challenges. Surgeons must control their natural hand tremor ($\sim$180 $\mu m$ in amplitude) \cite{hand_tremor_amplitude}, which is comparable to the thickness of the fragile retina ($\sim$200 $\mu m$) \cite{retinal_thickness}. Additionally, they face difficulties with depth perception during needle insertion. Moreover, maintaining precise needle-tip position during drug infusion for extended periods poses a risk of damaging the retina due to uncontrollable hand tremor \cite{first_human_subretinal_injection}. Consequently, unassisted SI pushes surgeons to their physiological limits \cite{first_human_subretinal_injection}.

To address the challenges of SI, previous studies have introduced robotic assistance and intraoperative optical coherence tomography (iOCT) for depth guidance. iOCT is an imaging modality that provides cross-sectional (B-scan) or volumetric (C-scan) views of the surgical workspace, providing depth perception during needle navigation and insertion. Using such systems, prior works have demonstrated robot-assisted SI under teleoperated control by a surgeon \cite{first_human_subretinal_injection}. More recently, additional efforts have been made toward automating SI. For example, \cite{mach2022oct} and \cite{shervin_ioct} developed workflows that allowed surgeons to select a goal waypoint in the iOCT view. The robot then navigated to the selected waypoint below the retina to accomplish needle insertion. However, they relied on the limited pure translational motion of the robot without enforcing safety-related kinematic constraints such as the remote-center-of-motion (RCM) constraint, which is necessary to avoid causing extraneous rotation on the patient's eye.

While there has been promising progress toward automation, significant challenges remain. This paper focuses on automating SI and thus, we highlight the limitations of prior works in this area, specifically \cite{mach2022oct} and \cite{shervin_ioct}. Firstly, these works lacked real-time capabilities due to reliance on slow volumetric C-scans. For instance, the state-of-the-art Leica iOCT system in \cite{shervin_ioct} required 7.69 seconds to scan a 2.5mm x 2.5mm (100 B-scans) square patch of the retina. Such scanning speed is inadequate considering the potential occurrence of involuntary patient motion and cyclic oscillations caused by patient breathing during that time period \cite{first_human_subretinal_injection}. While relying on smaller C-scans or a single B-scan is possible \cite{mach2022oct}, the resulting scanning region may become too limited, leading to the risk of the surgical instrument or the target tissue leaving the viewable field due to patient motion. Secondly, prior works have focused exclusively on utilizing iOCT views, neglecting the information in the microscope's global and intuitive color view. The microscope view provides important and familiar information to surgeons and should be utilized to identify affected regions (e.g., areas of bleeding). This is not possible with iOCT as it is limited to grayscale depth information. Lastly, prior works have not accounted for the important remote-center-of-motion (RCM) constraint when designing their workflows. They have instead relied only on the pure translational motion of the robot. It is crucial to enforce realistic surgical constraint to enable safe surgical workflows.

\begin{figure*}
        \centering
        \includegraphics[width = \textwidth]{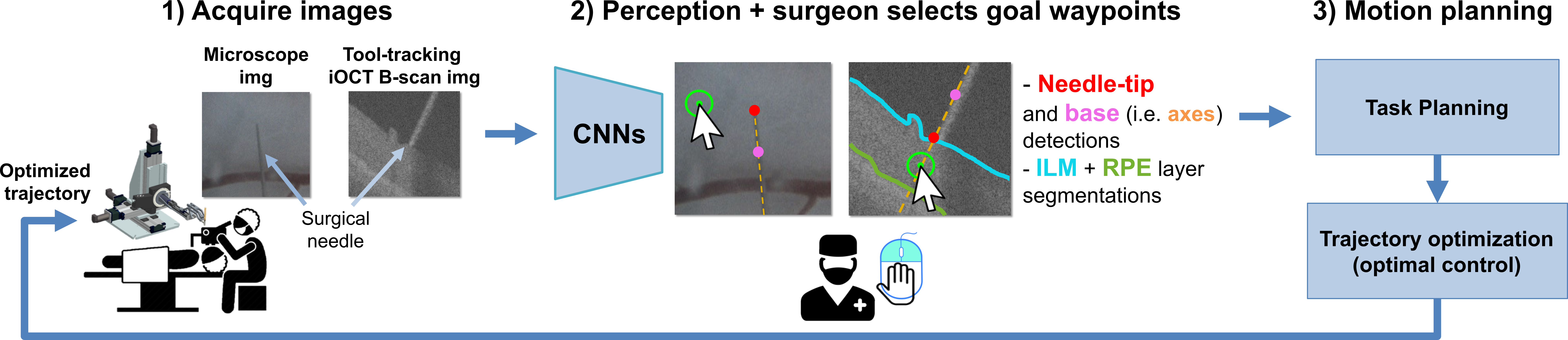}
        \caption{\small High-level workflow; 1) the microscope image and the iOCT B-scan images are acquired. 2) Three CNNs provide needle-tip detections and retinal layer segmentatoins required for task execution. The surgeon provides two waypoint goals in the microscope and B-scan images via two mouse-clicks. Note that the waypoint goal is first specified in the microscope image, and later on in the B-scan image right before the needle insertion step. 3) Based on these information, the relevant task and motion is planned, and an optimized trajectory is sent to the robot for trajectory-tracking.}
        \label{fig:workflow}
\end{figure*}

In this paper, we address the above limitations by first developing a custom imaging system that integrates the microscope and iOCT in a real-time manner. Unlike prior works which relied on slow volume scans \cite{shervin_ioct} or fixed cross-sectional scans \cite{mach2022oct}, we consider a hybrid approach where a small iOCT scanning region \emph{dynamically} tracks the surgical instrument. This is achieved by detecting the surgical instrument axis in the microscope image, and using this information to generate a B-scan aligned with the instrument axis. Even if the surgical tool moves, the B-scan automatically tracks the tool, thereby enabling real-time depth feedback between the needle and the underlying retinal tissue. Our system is also capable of providing C-scan volumes of any size while tracking the tool, but for simplicity we utilize B-scans to demonstrate a practical SI application. Ultimately, by combining microscopy and iOCT imaging, our system enables real-time global and local awareness of the surgical workspace, offering both intuitive color and precise depth information. This contribution addresses the first two limitations mentioned above. We tackle the third limitation by employing this system to design a real-time workflow incorporating the RCM constraint, ensuring patient safety. Our contributions include:

\begin{enumerate}[nolistsep]

\item Designing a system and workflow for real-time autonomous SI that utilizes microscope and tool-axis-aligned B-scan images, with the B-scan dynamically tracking the tool axis to provide real-time depth feedback between the needle and the retina. RCM constraint is enforced to ensure patient safety.

\item Outlining a strategy for calibrating the microscope and iOCT to generate tool-axis-aligned B-scans.

\item Validating the system via 30 successful autonomous trials on three cadaveric pig eyes, achieving mean needle insertion accuracy of $26 \pm 12 \mu m$ to various subretinal goals and requiring a mean duration of  $55 \pm 10.8$ seconds. Preliminary comparisons to a human operator in robot-assisted mode demonstrate the improved safety of our system during needle-tissue interactions.

\end{enumerate}

\section{Related Works}

    iOCT has been used in several robot-assisted surgical applications, including corneal keratoplasty \cite{keller2020optical}, vein cannulation \cite{gerber_vein_cannulation}, and subretinal injection \cite{shervin_ioct} \cite{mach2022oct}. However, the common assumption across these works was that the iOCT (either used in C-scan and B-scan mode) was fixed on a predefined region-of-interest (ROI). Given the speed limitations of acquiring C-scans or the limited view of B-scans, the proposed systems and workflows may not extend to dynamic settings where patient motion and instrument deformation are  present, and thus dynamic update of the iOCT ROI is necessary. Our present work addresses these limitations by implementing a tool-tracking iOCT system, and introducing the microscope view for a global and intuitive view of the surgical workspace.
        
    We also highlight a closely related work \cite{4d_ioct}, which demonstrated real-time ROI update of the OCT B-scan or C-scan view by detecting a bounding box of the surgical tool from top-down spectrally encoded reflectometry (SER) images, in a similar fashion to our system. Our work, however, implements the calibration using microscope images and uses a different calibration approach. Also, their work primarily focused on developing an imaging system without robotic integration or experimental validation using animal tissues. For a general comprehensive overview of technological innovations in retinal surgery, see \cite{robotic_surgery_review}.

\section{Problem Formulation} \label{problem_formulation}
\begin{figure}
        \centering
        \includegraphics[width = \columnwidth]{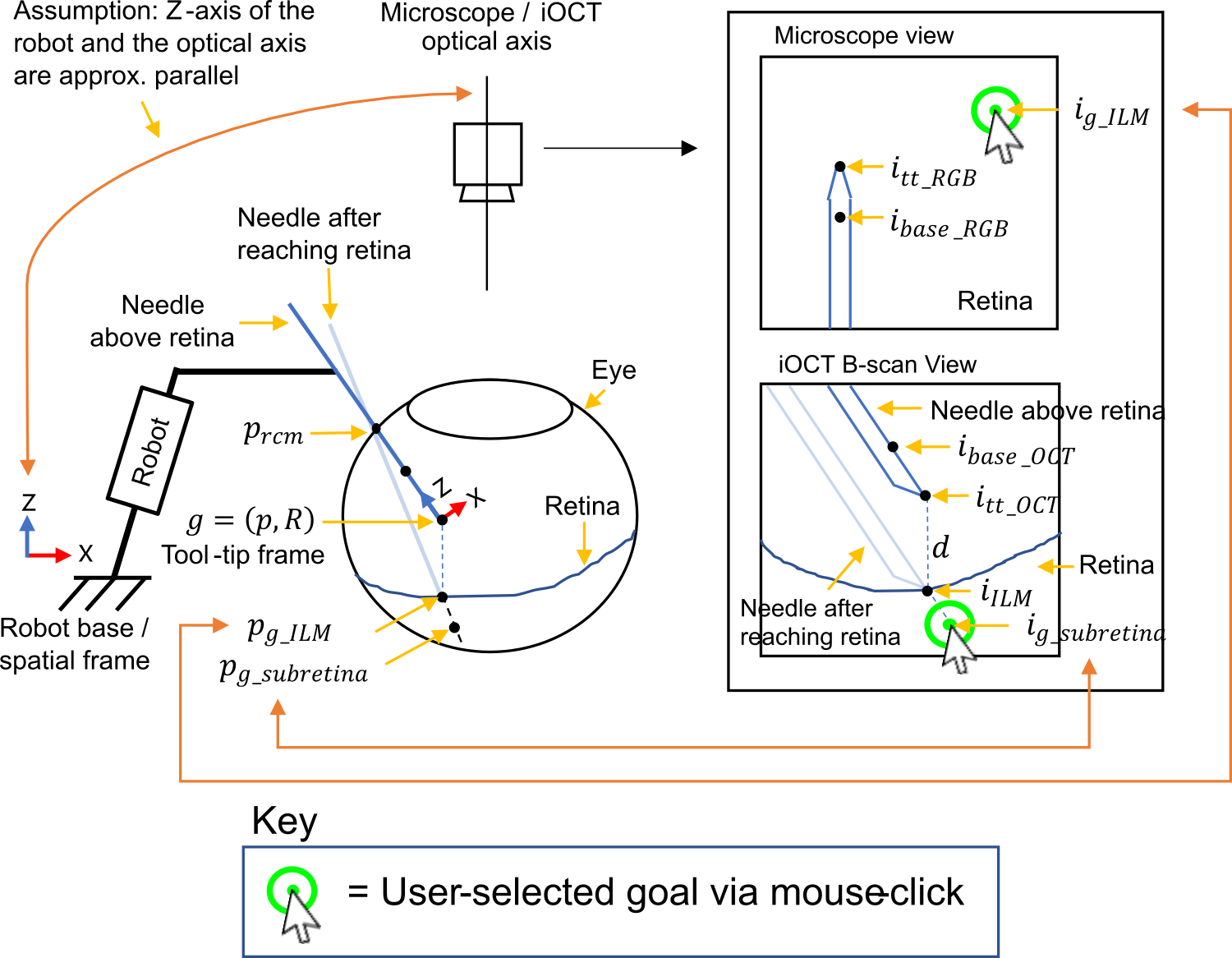}
        \caption{\small Key variables used are shown. Double arrows are shown to indicate that the pixel goals $i_{g\_ILM}$ and $i_{g\_subretina}$ correspond to $p_{g\_ILM}$ and $p_{g\_subretina}$ respectively in euclidean space.}
        \label{fig:variables_illustrated}
\end{figure}

Consider a robotic manipulator with a surgical tool attached at its end-effector, as illustrated in Fig. \ref{fig:variables_illustrated}. The key variables mentioned throughout this section are also illustrated in Fig. \ref{fig:variables_illustrated}. We define the robot states as $x=(g, V)$, where $g = (p, R) \in SE(3)$ and $V = (v, \omega) \in \mathbb{R}^6$. $p\in\mathbb{R}^3$ denotes the tool-tip position, $R\in SO(3)$ the orientation, and $v\in\mathbb{R}^3$ and $\omega\in\mathbb{R}^3$ the tool-tip-frame translational and angular velocity. Let the robot occupy a region $A(q)\subset \mathcal W$ in the workspace $\mathcal W\subset \mathbb{R}^3$, where $q$ denotes the robot joint angles.

The tool-tip state $x$ is fully-observable using high-precision motor encoders and precise knowledge of the robot forward kinematics. A given desired tool-tip velocity can be mapped to robot joint velocity to actuate the robot. Additionally, the system includes a monocular microscope camera  generating top-down observations of the surgical environment $o_{RGB}(t)\in \mathcal I$ from space of images $\mathcal I$ at a given time $t$. A co-axially mounted OCT generates B-scan images $o_{OCT}(t) \in \mathcal I$. The B-scan plane dynamically tracks the tool such that the scanning plane is always aligned with the tool axis, thereby providing depth feedback between the needle and the underlying retina at all times.

Initially, the surgeon manually introduces the surgical tool into the eye through a sclera entry point, $p_{rcm}\in\mathbb{R}^3$, which is recorded at the time of entry. The sclera point should remain fixed after each entry to avoid unsafe forces exerted on the sclera tissue. Once the tool-tip is within the microscope's view, its key points are detected in the microscope and B-scan views using two CNNs. Specifically, the detected tool-tip and its base are denoted as $i_{tt\_RGB}, \ i_{base\_RGB} \in o_{RGB}(t)$ in the microscope image  and $i_{tt\_OCT}, \  i_{base\_OCT} \in o_{OCT}(t)$ in the B-scan image. These detected points also define the axis of the tools in the respective images (Fig. \ref{fig:workflow}). Simultaneously, the ILM and RPE layers are segmented using another CNN, generating binary segmentation masks $I_{ILM}$ and $I_{RPE}$. Using the detected tool-tip $i_{tt\_OCT}$ and the segmented ILM layer $I_{ILM}$, the projection of the tool-tip to the ILM layer below (i.e. along the same B-scan image column index) is computed and denoted $i_{ILM}$ in the B-scan view. 
    
Given this setup, the surgeon selects a 2D pixel goal $i_{g\_ILM} \in o_{RGB}(t)$ via a mouse-click in the top-down microscope image. This goal denotes the desired waypoint through which the needle is introduced into the retina from the microscope view. It also corresponds to a 3D euclidean point $p_{g\_ILM} \in \mathcal{W}$ on the retina's surface w.r.t the robot's spatial frame. We seek to reach $p_{g\_ILM}$, however, its exact location is unknown. Instead of estimating $p_{g\_ILM}$ directly, we propose to reach it approximately by employing a specific visual-servoing strategy (Section \ref{technical_approach}). After $p_{g\_ILM}$ is reached using this strategy, the surgeon specifies another goal waypoint $i_{g\_subretina} \in o_{OCT}(t)$ along the axis of the needle and below the retina in the B-scan view via a mouse-click. Note that at this point in time, the needle is placed on the retinal surface, as illustrated in Fig. \ref{fig:variables_illustrated} (labelled as ``Needle after reaching retina''). The goal $i_{g\_subretina}$ corresponds to a 3D euclidean point $p_{g\_subretina} \in \mathcal{W}$ defined w.r.t the robot spatial frame. This subretinal goal is the final drug-delivery site. Since the iOCT B-scan is always aligned with the needle-axis, $p_{g\_subretina}$ can be reached by simply inserting it along its axis.

In summary, the objective is to navigate the needle-tip to two sequential goals: $p_{g\_ILM}$ and then $p_{g\_subretina}$, given the user-clicked goals $i_{g\_ILM}$ and $i_{g\_subretina}$ respectively. We thus consider the following two problems:

\begin{enumerate}
\item \emph{Navigating the needle above the retinal surface}: navigate the needle-tip to the desired needle insertion point on the retinal surface $p_{g\_ILM}$ given the clicked goal $i_{g\_ILM}$.

\item \emph{Needle insertion}: insert the needle along its axis to reach the goal insertion waypoint $p_{g\_subretina}$ given the clicked goal $i_{g\_subretina}$.
\end{enumerate}

The objective is to autonomously perform the above tasks while relying on monocular top-down images, tool-axis aligned B-scan images, and three CNNs automatically providing detections and segmentations necessary for task autonomy. Additionally, kinematic constraints concerned with the safety of the surgery must be satisfied while ensuring smooth robot motion.

\section{Technical Approach} \label{technical_approach}

\subsection{Navigation above the retina} \label{nav_section}

The first step of the navigation procedure is positioning the needle-tip at the first desired waypoint $p_{g\_ILM}$ given the user-selected goal $i_{g\_ILM}$. Recall that $i_{g\_ILM}$ is only a 2D pixel goal. Therefore, the corresponding 3D location $p_{g\_ILM}$ is unknown. One possible approach may be to estimate $p_{g\_ILM}$ directly using images and/or iOCT. However, this can be challenging in retinal surgery where unknown distortion is present. Instead, we propose a straightforward navigation procedure in which $p_{g\_ILM}$ can be approximately reached without directly estimating its 3D position. At a high level, this procedure first consists of aligning the needle-tip with the clicked goal $i_{g\_ILM}$ from the top-down microscope view via 2D planar motion. Then, the needle is simply lowered towards the retina while relying on the B-scan for depth feedback. The specific procedure is as follows:

\begin{enumerate}[label=Step \arabic*: , leftmargin=*]
    \item Align the needle-tip with the clicked goal $i_{g\_ILM}$ via 2D visual-servoing i.e. via actutation \emph{only} along the robot's spatial XY plane (the robot's spatial XY frame is shown in Fig. \ref{fig:variables_illustrated}). This step effectively aligns the needle-tip with the clicked goal pixel $i_{g\_ILM}$ from the top-down microscope view.
    
    \item Lower the needle towards the retinal surface via incremental motion along the robot's spatial Z-axis. This step moves the needle-tip closer to the retinal surface in the B-scan view, while mostly keeping the needle-tip aligned with the clicked goal $i_{g\_ILM}$ from the microscope view. 

    \item Repeat the above two steps until $i_{ILM}$ (Fig. \ref{fig:variables_illustrated}) is reached in the B-scan view.
\end{enumerate}

The underlying assumption here is that the optical axis of the microscope and the robot's spatial Z-axis are approximately parallel, as illustrated in Fig. \ref{fig:variables_illustrated}. Therefore, during 2D planar motion (step 1), the observed motion of the needle-tip is a corresponding planar motion in the microscope image. During the needle lowering step (step 2), the observed motion is a corresponding needle-lowering motion in the B-scan view. However, during step 2, the tool-tip may deviate from the clicked goal in the microscope view, since the optical axis and the robot's spatial Z-axis are only approximately parallel. Therefore, steps 1 and 2 must be repeated (step 3) whenever a pixel error above a small threshold is observed to realign the needle-tip with the clicked goal. In our experiments, we chose this threshold to be 1 pixel. Also, in order to avoid potential collision between the needle and the retina, we enforced a safety factor by offsetting $i_{ILM}$ by 30$\mu m$ above the retinal surface. Ultimately, this iterative procedure enables accurate needle-tip placement anywhere on the retinal surface. Note that to achieve step 1, calibration between the robot and the microscope is necessary, which is described in Section \ref{vs_section}.


\subsection{Needle insertion}

Once the needle-tip is approximately placed on $p_{g\_ILM}$ (Section \ref{nav_section}), the surgeon specifies another goal waypoint $i_{g\_subretina}$ in the B-scan view via a mouse click. The desired insertion distance is obtained as follows:

\begin{align}
    d_{insertion} = \lVert i_{g\_subretina} - i_{tt\_OCT} \rVert^2,  
\end{align}
where $d_{insertion}$ is converted to microns using a conversion factor (Table \ref{tab:table_parameters}). The needle is then simply inserted along its axis to reach $p_{g\_subretina}$. 

\subsection{Microscope-OCT Calibration}
\label{oct_section}
The microscope-OCT setup combines a 100 kHz swept source OCT system~\cite{wei2019analysis} with a microscope for simultaneous OCT and microsopic imaging. A charged-coupled device (CCD) is added to the OCT system to capture microscopic images~\cite{wei2022region}. Both the OCT and the CCD share the same objective lens, ensuring precise alignment and consistent working distance. A short-pass dichroic mirror with a 650 nm cutting-off wavelength splits the light into reflection and transmission. It reflects near infrared light back to the OCT system while transmitting visible light to the CCD for microscopic imaging. The transmitted visible light is focused onto the CCD by an imaging lens, and a short-pass filter is employed to reduce near infrared noise. Two galvo mirrors are utilized to tilt the collimated beam from the fiber collimator, and thus control the OCT scanning position. This integrated microscope-OCT setup facilitates a comprehensive OCT and microscopic visualization.

\begin{figure}
    \centering
    \includegraphics[width = \columnwidth]{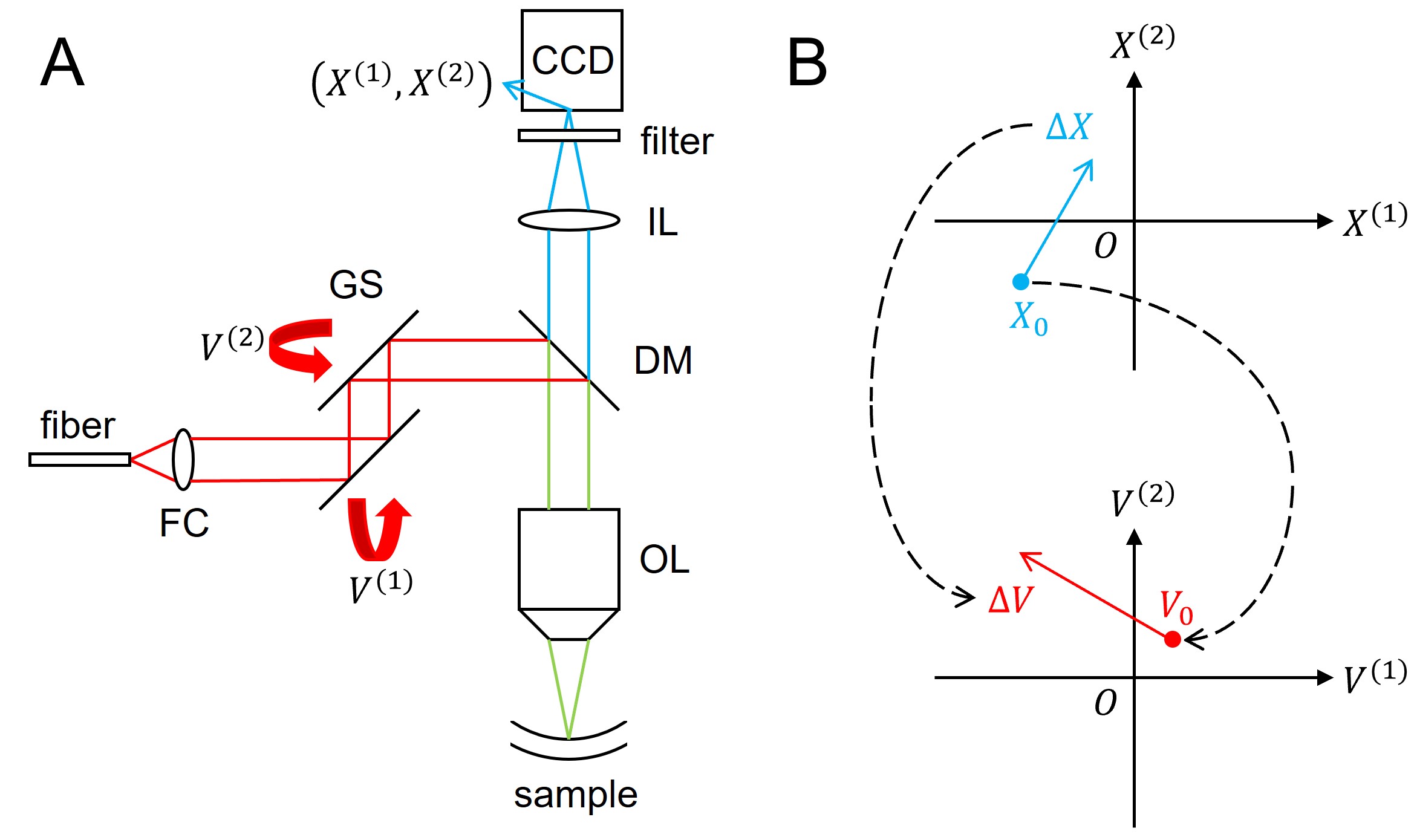}
    \caption{(A)~Microscope-OCT system setup. FC, fiber collimator; GS, galvo scanners; DM, dichroic mirror; IL, imaging lens; OL, objective lens. (B)~The mapping between the laser scanning position and the applied voltage.}
    \label{fig:oct_setup}
\end{figure}

The laser scanning position in a microscopic image is determined by the rotation angles of two orthogonal galvo mirrors, which are controlled through voltages. We assume a linear relationship between the laser scanning position and the applied voltages, given by the equation:
\begin{equation}
    X = R V + T,
    \label{eq:oct_01}
\end{equation}
where $X = \begin{bmatrix}
    X^{(1)} & X^{(2)}
\end{bmatrix}^\intercal \in \mathbb{R}^2$ represents the laser scanning position in the microscopic image, $V = \begin{bmatrix}
    V^{(1)} & V^{(2)}
\end{bmatrix}^\intercal \in \mathbb{R}^2$ corresponds to the voltages applied to the two galvo mirrors, $R \in \mathbb{R}^{2 \times 2}$ and $T \in \mathbb{R}^2$ are linear parameters that convert the applied voltages to the corresponding laser scanning position in the microscopic image. The mapping between the laser scanning position and the applied voltage is shown in Fig.~\ref{fig:oct_setup}(b). 

The linear parameters $R$ and $T$ need to be calibrated. We use a laser viewing card to visualize and locate the laser scanning position in the microscopic image. We record a set of laser scanning positions $\{X_i|i \in \{1,\dots,N\}\}$ when applying voltages according to a predefined voltage set $\{V_i|i \in \{1,\dots,N\}\}$. The linear parameters $R$ and $T$ are calibrated by minimizing the least square error:
\begin{equation}
    \hat{R}, \hat{T} = \operatorname*{argmin}_{R,T}\sum_{i=1}^N ||R V_i + T - X_i||_2^2,
    \label{eq:oct_02}
\end{equation}
where $||\cdot||_2$ refers to Euclidean norm. It can be derived that:
\begin{equation}
    \hat{R}^\intercal = (\mathbb{V} \mathbb{V}^\intercal)^{-1} \mathbb{V} \mathbb{X}^\intercal,
    \label{eq:oct_03}
\end{equation}
\begin{equation}
    \hat{T} = \bar{X} - \hat{R} \bar{V},
    \label{eq:oct_04}
\end{equation}
where $\mathbb{V} \in \mathbb{R}^{2 \times N}$ and $\mathbb{V} \in \mathbb{R}^{2 \times N}$ are matrices that are expressed as:
\begin{equation}
    \mathbb{V} = \begin{bmatrix}
    V_1-\bar{V} & V_2-\bar{V} & \cdots & V_N-\bar{V}
    \end{bmatrix},
    \label{eq:oct_05}
\end{equation}
\begin{equation}
    \mathbb{X} = \begin{bmatrix}
    X_1-\bar{X} & X_2-\bar{X} & \cdots & X_N-\bar{X}
    \end{bmatrix},
    \label{eq:oct_06}
\end{equation}
and $\bar{V} = \frac{1}{N} \sum_{i=1}^{N} V_i$, $\bar{X} = \frac{1}{N} \sum_{i=1}^{N} X_i$.

To generate OCT B-scan that is aligned with the needle axis in the microscopic image, we need the needle tip position, needle orientation and a predefined scanning length in the microscopic image frame. The central voltage is determined by:
\begin{equation}
    V_0 = \hat{R}^{-1} (X_0 - \hat{T}),
    \label{eq:oct_07}
\end{equation}
where $X_0 \in \mathbb{R}^2$ is the needle tip position in the microscopic image. The voltages are determined on the tangent space at $V_0$ through the equation:
\begin{equation}
    \Delta V = \hat{R}^{-1} \Delta X,
    \label{eq:oct_08}
\end{equation}
where $\Delta V$ is the tangent vector at $V_0$ describing the voltage amplitude and the voltage angle, and $\Delta X$ is the tangent vector at $X_0$ describing the needle orientation and the predefined scanning length. Therefore, to generate a scanning cross section along the needle axis and centered at the needle tip that is described by:
\begin{equation}
    X(t) = X_0 + t \Delta X,
    \label{eq:oct_09}
\end{equation}
where $t \in (-1,1)$ is a normalized time parameter, we should control the voltage according to:
\begin{equation}
    V(t) = V_0 + t \Delta V,
    \label{eq:oct_10}
\end{equation}
where $V_0$ and $\Delta V$ are determined by Eq.~\ref{eq:oct_07} and Eq.~\ref{eq:oct_08}, respectively. 




\subsection{Real-Time Hand-Eye Calibration and Visual Servoing} \label{vs_section}

In order to navigate the needle-tip to the clicked goal $i_{g\_ILM}$ (i.e. during step 1 in Section \ref{nav_section}), the calibration parameters between the robot and the microscope must be ascertained. To avoid complex calibration procedure such as in \cite{mach2022oct}, we choose a visual servoing strategy which relies on real-time iterative updates to the calibration matrix based on the observed robot motions in real-time \cite{jagersand_vs}. This effectively enables real-time adaptation to the changing intrinsic and extrinsic properties of the camera, enabling the surgeon to change the microscope position, magnification, or add distortive optics (e.g. BIOME or contact-lenses) during the procedure, without needing to perform a calibration procedure repeatedly.

Since the robot's spatial Z-axis and the camera's optical axis are approximately parallel (Section \ref{nav_section}), the calibration is only performed between the robot's spatial XY plane and the image plane. To simplify the notation, we introduce a variable $\bar{p} = Sp \in \mathbb{R}^2$, which simply denotes the XY components of the surgical tool-tip position, where the selector matrix $S \in \mathbb{R}^{2 \cross 3}$ picks out the first two elements of the vector it operates on. $S$ is defined as:
\begin{align}
S = \begin{bmatrix}
1 & 0 & 0 \\
0 & 1 & 0 
\end{bmatrix}.
\end{align}

Consider an unknown function $K:\mathbb{R}^2 \rightarrow \mathbb{R}^2$ which converts the tool-tip XY position to its corresponding image coordinates. $K$ implicitly contains the intrinsic and extrinsic parameters of the camera. In other words,
\begin{equation}
i_{tt\_RGB} = K(\bar{p}).
\end{equation}

We may approximate the unknown $K$ using the first-order Taylor series approximation:
\begin{align}
 K(\bar{p}^{k+1}) &\approx K(\bar{p}^k) + J_{calib}(\bar{p}^k)(\bar{p}^{k+1}-\bar{p}^k) \\ 
\Delta i_{tt\_RGB}^k & \approx J_{calib}(\bar{p}^k)(\Delta \bar{p}^k) \label{eq:jacobian_eq}
\end{align}
where $J_{calib}$ is a Jacobian matrix that relates the change in tool-tip XY position $\Delta \bar{p}^k$ to the corresponding change in image coordinates $\Delta i_{tt\_RGB}^k$. $k$ denotes the iteration step, since the Jacobian is iteratively updated to adapt to observed motions. Specifically, the Jacobian is recalculated whenever a significant change $\Delta i_{tt\_RGB}^k$ and $\Delta \bar{p}^k$ is observed. In our experiments, we updated the $J_{calib}$ when $\Delta i_{tt\_RGB}^k > 8$ pixels and $\Delta \bar{p}^k > 20 \mu m$) was observed.

Borrowing from \cite{jagersand_vs}, we use an online update rule to estimate the Jacobian in real-time based on the observed robot motions. The method utilizes Broyden's update formula to estimate the Jacobian, given as:
\begin{equation} \label{eq:broyden}
    J_{calib}^{k+1} = J_{calib}^{k} + \beta \frac{\Delta i_{tt\_RGB}^k - J_{calib}^k \Delta \bar{p}^k}{(\Delta \bar{p}^k )^T (\Delta \bar{p}^k)} (\Delta \bar{p}^k)^T,
\end{equation}
where $0 \leq \beta \leq 1$ is the step size for updating the Jacobian \cite{Broydens_method}. We chose $\beta = 0.5$. This is an iterative approach where the Jacobian is initialized as an arbitrary non-singular matrix (e.g. an identity matrix) and after several updates it converges to the true Jacobian. To use the Jacobian to guide the needle to $i_{g\_ILM}$, we reformulate Eq. \ref{eq:jacobian_eq} as
\begin{equation} \label{eq:visual_servoing}
    \Delta \bar{p}_{desired}= J_{calib}(\bar{p}^k)^{-1} \Delta i_{desired},
\end{equation}
where $\Delta i_{desired} = i_{g\_ILM} - i_{tt\_RGB}$ is the desired motion vector in image coordinates, and $\bar{p}_{desired}$ is the desired change in tool-tip position to align the needle-tip with the clicked goal.

Using $\Delta \bar{p}_{desired}$, the desired waypoint to reach can be given as:

\begin{equation} \label{eq:visual_servoing}
    p_{desired}= p + \begin{bmatrix}
           \Delta \bar{p}_{desired} \\
          0 \\ 
           \end{bmatrix}.
\end{equation}
where the motion along robot's spatial Z-axis is zero. To be clear, $p_{desired} \neq p_{g\_ILM}$. $p_{desired}$ can be considered as an intermediate waypoint that is constantly updated (e.g. every time the Jacobian is updated), such that the tool-tip will be eventually aligned with $i_{g\_ILM}$ in the microscope view.


\subsection{Optimal Control Formulation} \label{optimal_control}

Once a desired goal waypoint $p_{desired}$  is determined (Section \ref{vs_section}) it is used by an optimal control framework to generate an optimal trajectory to the goal. The formulation and implementation follows our previous work \cite{eye_surgery_imitation_learning}.

\subsection{Network Detail} \label{network_training}
\begin{figure}
        \centering
        \includegraphics[width = \columnwidth]{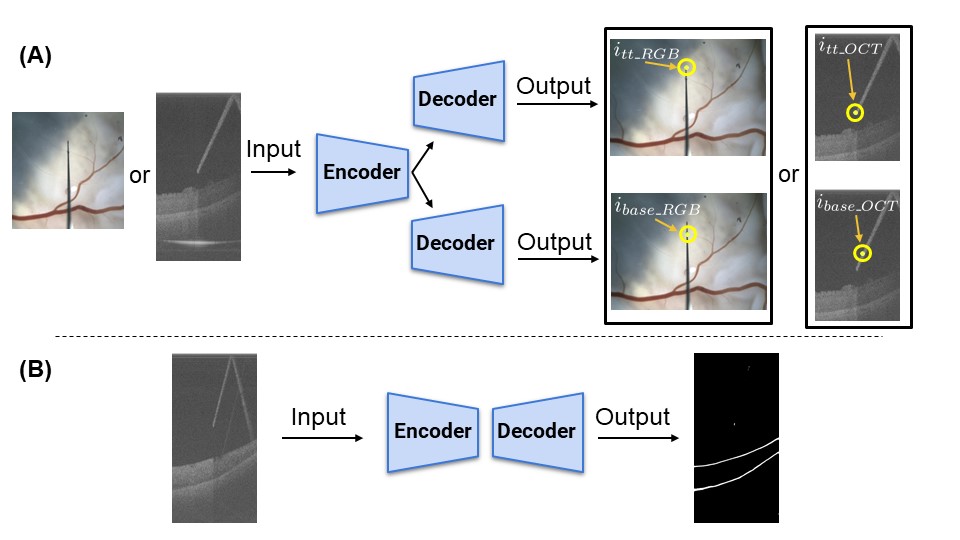 }
        \caption{\small Network architectures: (A) Two networks are trained to detect the needle tip and its base (thus defining its axis), one for microscope and another for iOCT images. (B) A third network is trained to the ILM and RPE layer segmentations}
        \label{fig:network_structure}
\end{figure}

Two CNNs are implemented for the tool-tip and its base predictions (Fig. \ref{fig:network_structure}A). One network is used for the microscope images and another network for the B-scan images. Both networks are identical and adopt a U-Net-like \cite{ronneberger2015u} architecture, using Resnet-18 \cite{he2016deep} as a backbone. After the image is encoded into feature vectors, two decoders are used to predict the tool-tip and its base respectively. The output sizes are identical to the input sizes ($480 \times 640 \times 3$ and $1024 \times 512$ for the microscope and B-scan images respectively). In order to enforce consistency, the distance between the tool-tip and its base are set to be 50 and 100 pixels in the microscope and B-scan images respectively. 

We implement a third CNN to segment the ILM  and the RPE layer in the B-scan image, as shown in Fig. \ref{fig:network_structure}B. This segmentation network is identical as before, but the decoder now outputs three channels, predicting the background, ILM layer, and the RPE layer respectively.

To train the networks, cross-entropy loss was used for the tool-tip and its base predictions and the retinal layers segmentation. For the tool-tip and its base predictions, an additional MSE loss was used to enforce consistent distance spacing between the predicted points (i.e. 50 or 100 pixels). Also, to balance the errors among the three labels during segmentation training, we set the weights for the background, ILM, and RPE layer to be 0.001, 0.4995, and 0.4995 respectively. 2000 microscope images and 1050 B-scan images were used for training. Adam optimizer with a learning rate of 0.0003 was used.

\subsection{Hardware}

Our system consisted of three computers: one computer for controlling the surgical robot, another for acquiring images from the iOCT hardware, and a GPU workstation for running the CNNs and displaying the GUI. Our system was setup was not optimized, and an improved setup should consider integrating the hardware and software into a single powerful computer. The data between the computers were communicated via ethernet cables. The greatest bottleneck was sending the iOCT B-scan images to the GPU workstation, and due to latency the B-scan images were received at 11Hz. The camera was directly connected to the GPU workstation and its images were acquired at 30Hz. The inference speed of all the networks described in \ref{network_training} exceeded 40Hz, and they were deployed on an NVIDIA RTX 3090 GPU.

\section{Experiments} \label{experiments_section}
As shown in Fig. \ref{fig:intro_fig}B, the experimental setup consists of the Steady Hand Eye Robot (SHER) \cite{Steadyhand2}, a silver-coated glass pipette attached at the robot handle ($30\mu m$ tip diameter), a microscope-integrated OCT, and an open-sky pig eye filled with vitreous.

\begin{figure*}
        \centering
        \includegraphics[width = \textwidth]{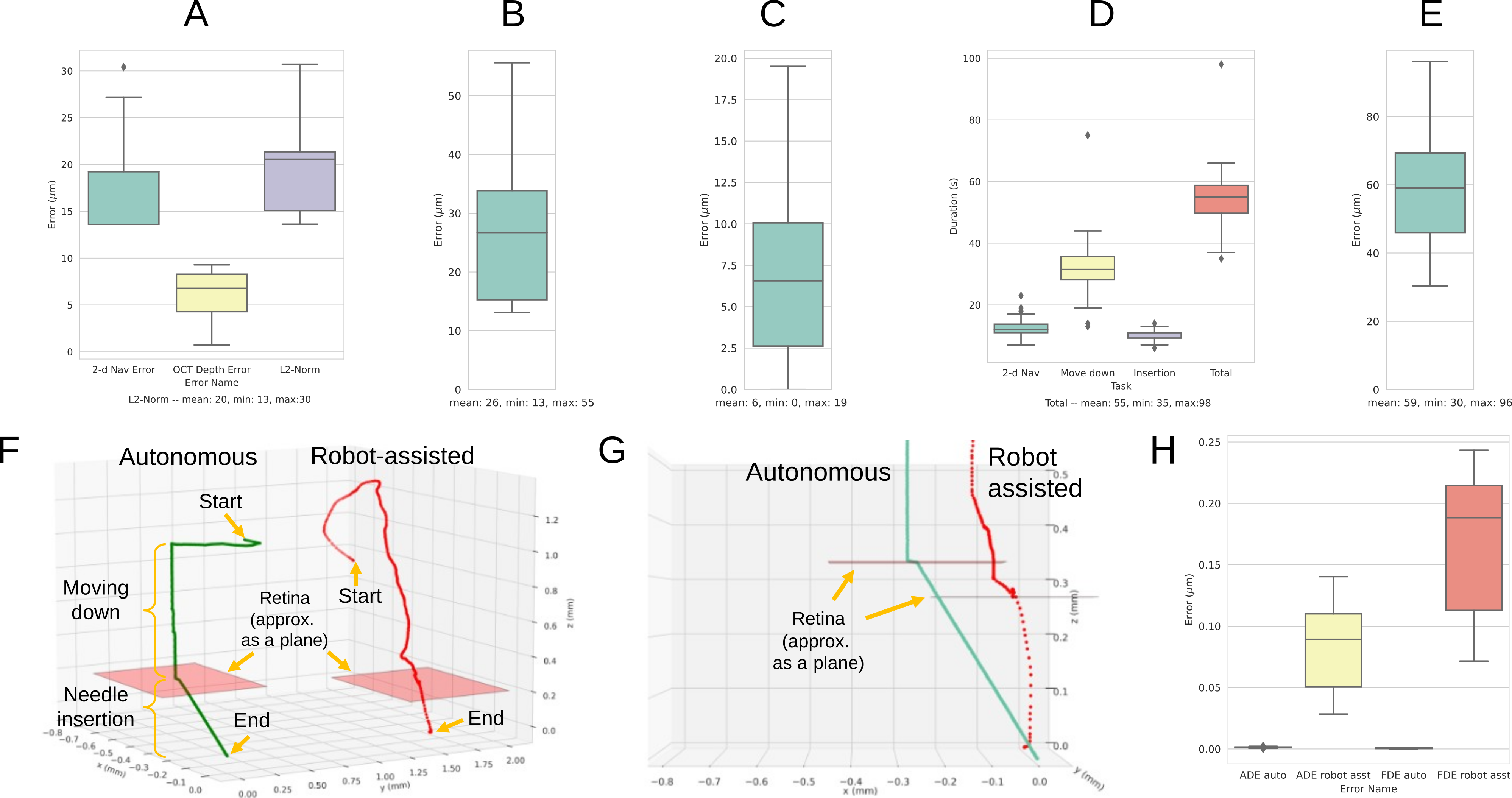}
        \caption{\small Experimental metrics are shown (see Section \ref{experiments_section}): (A) navigation error on the retinal surface goal $p_{g\_ILM}$ (B) needle-insertion error at the drug delivery site $p_{g\_subretina}$ (C) RCM error (D) total duration of the surgery by task (E) 2D navigation error to the clicked goal from the microscope view during robot-assisted mode (F) qualitative comparison between autonomous and robot-assisted modes (G) same comparison from a close-up side view during needle insertion (H) comparing the deviation of the needle-tip from the insertion axis during needle insertion; ADE (average displacement error), FDE (final displacement error).}
        \label{fig:result_plots}
\end{figure*}

We validated our system through 30 autonomous subretinal injection trials on 3 open-sky pig eyes (Fig. \ref{fig:intro_fig}B). For each eye, 10 trials were performed. The experimental procedure follows the description provided in Section \ref{problem_formulation}. We evaluate our system based on the following metrics:


\begin{table}[hbt!]
\begin{center}
\caption{Conversion factors for converting pixel distance to microns}
\label{tab:table_parameters}
\begin{tabular}{|c|c|}
\hline
Image Type              & Conversion factor ($\mu m$ / pixel) \\ \hline
Microscope image      & 13.6                                             \\ \hline
B-scan (along img height) & 2.6                                              \\ \hline
B-scan (along img width)  & 5.3                                              \\ \hline
B-scan (between B-scan slices)  & 13.6                                             \\ \hline
\end{tabular}
\end{center}
\end{table}

\begin{enumerate}[wide, labelwidth=!, labelindent=0pt]

    \item \textbf{Navigation error on the retinal surface goal} (Fig. \ref{fig:result_plots}A): this metric measures how closely the needle-tip is placed on the retinal surface goal $p_{g\_ILM}$ after performing the navigation procedure described in Section \ref{nav_section}. Recall that we did not estimate of $p_{g\_ILM}$ directly. Therefore, we approximate the error by combining the 2D navigation error between the clicked goal ($i_{g\_ILM}$) and the needle-tip ($i_{tt\_RGB}$) in the microscope view and the depth error between the needle-tip ($i_{tt\_OCT}$) and the retinal goal ($i_{g\_ILM}$) in the B-scan view to obtain the navigation error in 3D. 
    
    Specifically, we compute the 2D navigation error in the microscope view using the following formula (2D Nav Error in Fig. \ref{fig:result_plots}A): $\lVert i_{g\_ILM} - i^*_{tt\_RGB}\rVert_2$, where $i^*_{tt\_RGB}$ is the ground-truth needle-tip pixel coordinate obtained via manual annotation. We compute the depth error using the B-scan view via the following formula (OCT Depth Error in Fig. \ref{fig:result_plots}A): $\lVert  \bar{S}^T i_{ILM} - \bar{S}^T i^*_{tt\_OCT} \rVert_2$, where $i^*_{tt\_OCT}$ is the ground-truth needle-tip pixel coordinate in the B-scan image obtained via manual annotation, and $\bar{S}$ is a selector vector that picks out the second element of the vector it operates on (i.e. the pixel index along the B-scan image height). Specifically, $\bar{S} = [0 \ 1] \in \mathbb{R}^{2}$. Finally, these two errors are combined using the L2-norm metric (L2-Norm in Fig. \ref{fig:result_plots}A). The computed pixel errors are converted into micrometers using the relevant conversion factors listed in Table \ref{tab:table_parameters}.     

    \item \textbf{Needle insertion error} (Fig. \ref{fig:result_plots}B): this metric measures how closely the needle-tip reaches the desired insertion goal below the retina $p_{g\_subretina}$ based on the acquired volume scans after needle insertion. The error is calculated using the following L2-norm metric: $\lVert [ i^T_{g\_subretina} \ gt\_slice ] - [ {i^{*T}_{tt\_OCT}} \ actual\_slice] \rVert_2$, where $gt\_slice$ is the ground-truth B-scan slice index which the needle is expected to land and $actual\_slice$ is the B-scan slice index which the needle actually lands after needle insertion. The computed voxel errors are converted to micrometers using the relevant conversion factors in Table \ref{tab:table_parameters}. Note that a volumetric scan was performed after needle insertion (i.e. multiple B-scans or ``slice" of images were collected) to compute this error. 

    \item \textbf{RCM error} (Fig. \ref{fig:result_plots}C): this metric measures how closely the needle-axis is aligned with the RCM point throughout the entire procedure, computed using robot kinematics data \cite{eye_surgery_imitation_learning}.

    \item \textbf{Task duration} (Fig. \ref{fig:result_plots}D): this metric measures the duration taken for each task and their total sum. We include the duration of the navigation procedure and the insertion procedure, excluding the time taken by the user to specify goal waypoints.
\end{enumerate} 

\section{Results and Discussion}

We provide more details on the results and the metrics provided in Section \ref{experiments_section}. During the 30 autonomous SI trials, the needle-tip could reach the desired retinal surface goal $p_{g\_ILM}$ with mean error of $20 \pm 6 \mu m$ w.r.t the L2-norm metric as shown in Fig. \ref{fig:result_plots}A. As shown in Fig. \ref{fig:result_plots}B, the subretinal goal $p_{g\_subretina}$ was reached with mean error of $26 \pm 12 \mu m$. Note that reaching the desired depth is the most critical requirement in terms of safety while reaching $p_{g\_subretina}$, to avoid potential damage to the retina, and the mean error along the depth dimension was $7 \pm 11 \mu m$. In comparison, such level of accuracy may be difficult to achieve for human surgeons, considering that the mean hand-tremor amplitude during retinal surgery is approximately 180$\mu m$ \cite{hand_tremor_amplitude}. Also, our navigation accuracy in reaching the subretinal goal was comparable to prior work of \cite{shervin_ioct}, which reported mean error of $32 \pm 4 \mu m$ and of \cite{mach2022oct}, which reported $50.4 \pm 29.8 \mu m$.  Throughout the entire procedure, the mean RCM error was kept low at $6 \pm 4 \mu m$ as shown in Fig. \ref{fig:result_plots}C. Finally, the mean total duration of the surgery was $55 \pm 10.8$ seconds as shown in  Fig. \ref{fig:result_plots}D.


 We also show a preliminary comparison to a human performing SI in robot-assisted mode. In robot-assisted mode, the human operator controlled the robot by placing the hand at the end-effector and modulating the gain of the motion using a foot pedal, and 10 trials were performed. The robot-assisted mode follows the control scheme originally developed in \cite{Steadyhand2}. Qualitatively, as shown in Fig. \ref{fig:result_plots}F, the autonomous trajectory was more stable and efficient.  Quantitatively, the robot-assisted trajectory was  less accurate in being able to reach the top-down clicked goal. Specifically, the mean navigation error for robot-assisted mode was $59 \pm 19 \mu m$ (Fig. \ref{fig:result_plots}E), while for autonomous mode it was $19 \pm  6\mu m$ (2D Nav Error in Fig. \ref{fig:result_plots}A) with $p = 4.15 \times 10^{-12}$ using a two-sided T-test. A closer side view comparison in Fig. \ref{fig:result_plots}G shows that, for the autonomous mode, the needle insertion trajectory was nearly perfect along the needle's axis. However, this is difficult to achieve in robot-assisted mode, since this constraint is difficult for humans to enforce by hand. We quantitatively show the deviation of the needle-tip trajectory from the originally intended insertion axis in Fig. \ref{fig:result_plots}H. Specifically, we consider a commonly-used average displacement (ADE) and final displacement error (FDE) metric for comparing trajectories. ADE computes the averaged L2-norm distance between the originally-intended insertion axis trajectory (i.e. straight-line trajectory along the needle's axis) and the executed trajectory, and FDE computes the L2-norm error between them only at the last waypoint of each trajectory.  Both ADE and FDE errors are near zero in autonomous mode. However, in robot-assisted mode, the errors are in the order of hundreds of micrometers. In summary, the autonomous mode is able to execute a more smooth motion, while keeping the needle trajectory constrained along its axis, thereby inflicting minimal damage on the retina during needle insertion.

 We also note that while all 30 autonomous trials were successful, 2 trials required human intervention. Intervention was necessary during the initial 2D navigation step above the retina due to perception error, when the CNN failed to detect the needle-tip position in the microscope image. When this occurred, the operator simply intervened and initialized the needle at a different location and the trial was resumed. This error did not lead to any damage on the retina, since the errors occurred during the navigation step above the retinal surface. To avoid such error, more efforts should be invested in improving the robustness of the CNN's predictions. However, tsuch work requires intensive labor for collecting diverse dataset and labelling them, which is out-of-scope in this work. We thus leave such improvement for future work. 
 
 Also, our system was demonstrated on a simplified open-sky eye filled with vitreous, which introduced some distortion. In a realistic setting, however, there will be other optical elements such as the patient's lens and cornea, which may introduce further distortion. Such distortions may affect our 2D visual-servoing strategy (Section \ref{vs_section}) but we provide explanation of how our approach may still work in heavier distortive settings. Specifically, our 2D visual-seroving strategy relies on a Jacobian matrix that assumes a linear relationship between $\Delta i_{tt\_RGB}$ and $\Delta \bar{p}$ (Section \ref{vs_section}). Generally, this relationship holds locally even in a distortive environment. Beacuse this Jacobian is updated when small local tool-tip motion is observed ($>8$ pixels and $>20\mu m$ in robot coordinates, see Setion \ref{vs_section}), as long as the linear relation generally holds within such local environment, our 2D navigation approach would still work. In practice, the observed phenomena in our open-sky eye setting was that as the needle-tip navigated to the clicked goal, the Jacobian was constantly updated such that it locally adpated to the distortive environment, which ultimately enabled precise navigation to the goal. Proving this system in a realistic eye setting will be left for future work.


 \section{Conclusion}
 We demonstrated a real-time autonomous system and workflow for subretinal injection. This was enabled by the global view provided by microscope images and dynamically-aligned B-scan images that tracked the needle axis for real-time depth feedback. Future work will consider extending this work to closed pig eye settings, and extending this framework to more challenging procedures (e.g. tissue grasping using forceps).

%

\bibliographystyle{IEEEtran}
\bibliography{bib}  

\begin{thebibliography}{10}
\providecommand{\url}[1]{#1}
\csname url@samestyle\endcsname
\providecommand{\newblock}{\relax}
\providecommand{\bibinfo}[2]{#2}
\providecommand{\BIBentrySTDinterwordspacing}{\spaceskip=0pt\relax}
\providecommand{\BIBentryALTinterwordstretchfactor}{4}
\providecommand{\BIBentryALTinterwordspacing}{\spaceskip=\fontdimen2\font plus
\BIBentryALTinterwordstretchfactor\fontdimen3\font minus
  \fontdimen4\font\relax}
\providecommand{\BIBforeignlanguage}[2]{{%
\expandafter\ifx\csname l@#1\endcsname\relax
\typeout{** WARNING: IEEEtran.bst: No hyphenation pattern has been}%
\typeout{** loaded for the language `#1'. Using the pattern for}%
\typeout{** the default language instead.}%
\else
\language=\csname l@#1\endcsname
\fi
#2}}
\providecommand{\BIBdecl}{\relax}
\BIBdecl

\bibitem{first_human_subretinal_injection}
J.~Cehajic-Kapetanovic, K.~Xue, T.~L. Edwards, T.~C. Meenink, M.~J. Beelen,
  G.~J. Naus, M.~D. de~Smet, and R.~E. MacLaren,
  ``\BIBforeignlanguage{en}{First-in-human robot-assisted subretinal drug
  delivery under local anesthesia},'' \emph{\BIBforeignlanguage{en}{Am. J.
  Ophthalmol.}}, vol. 237, pp. 104--113, May 2022.

\bibitem{hand_tremor_amplitude}
S.~Singh and C.~Riviere, ``Physiological tremor amplitude during retinal
  microsurgery,'' in \emph{Proceedings of the IEEE 28th Annual Northeast
  Bioengineering Conference (IEEE Cat. No.02CH37342)}, 2002, pp. 171--172.

\bibitem{retinal_thickness}
A.~Chan, J.~S. Duker, T.~H. Ko, J.~G. Fujimoto, and J.~S. Schuman,
  ``\BIBforeignlanguage{en}{Normal macular thickness measurements in healthy
  eyes using stratus optical coherence tomography},''
  \emph{\BIBforeignlanguage{en}{Arch. Ophthalmol.}}, vol. 124, no.~2, pp.
  193--198, Feb. 2006.

\bibitem{mach2022oct}
K.~Mach, S.~Wei, J.~W. Kim, A.~Martin-Gomez, P.~Zhang, J.~U. Kang, M.~A.
  Nasseri, P.~Gehlbach, N.~Navab, and I.~Iordachita, ``Oct-guided robotic
  subretinal needle injections: A deep learning-based registration approach,''
  in \emph{2022 IEEE International Conference on Bioinformatics and Biomedicine
  (BIBM)}.\hskip 1em plus 0.5em minus 0.4em\relax IEEE, 2022, pp. 781--786.

\bibitem{shervin_ioct}
S.~Dehghani, M.~Sommersperger, P.~Zhang, A.~Martin-Gomez, B.~Busam,
  P.~Gehlbach, N.~Navab, M.~A. Nasseri, and I.~Iordachita, ``Robotic navigation
  autonomy for subretinal injection via intelligent real-time virtual ioct
  volume slicing,'' 2023.

\bibitem{keller2020optical}
B.~Keller, M.~Draelos, K.~Zhou, R.~Qian, A.~N. Kuo, G.~Konidaris, K.~Hauser,
  and J.~A. Izatt, ``Optical coherence tomography-guided robotic ophthalmic
  microsurgery via reinforcement learning from demonstration,'' \emph{IEEE
  Transactions on Robotics}, 2020.

\bibitem{gerber_vein_cannulation}
M.~J. Gerber, J.-P. Hubschman, and T.-C. Tsao, ``Automated retinal vein
  cannulation on silicone phantoms using optical-coherence-tomography-guided
  robotic manipulations,'' \emph{IEEE/ASME Transactions on Mechatronics},
  vol.~26, no.~5, pp. 2758--2769, 2021.

\bibitem{4d_ioct}
\BIBentryALTinterwordspacing
E.~M. Tang, M.~T. El-Haddad, S.~N. Patel, and Y.~K. Tao, ``Automated
  instrument-tracking for 4d video-rate imaging of ophthalmic surgical
  maneuvers,'' \emph{Biomed. Opt. Express}, vol.~13, no.~3, pp. 1471--1484, Mar
  2022. [Online]. Available:
  \url{https://opg.optica.org/boe/abstract.cfm?URI=boe-13-3-1471}
\BIBentrySTDinterwordspacing

\bibitem{robotic_surgery_review}
I.~I. Iordachita, M.~D. De~Smet, G.~Naus, M.~Mitsuishi, and C.~N. Riviere,
  ``Robotic assistance for intraocular microsurgery: Challenges and
  perspectives,'' \emph{Proceedings of the IEEE}, vol. 110, no.~7, pp.
  893--908, 2022.

\bibitem{wei2019analysis}
S.~Wei, S.~Guo, and J.~U. Kang, ``Analysis and evaluation of bc-mode oct image
  visualization for microsurgery guidance,'' \emph{Biomedical optics express},
  vol.~10, no.~10, pp. 5268--5290, 2019.

\bibitem{wei2022region}
S.~Wei, J.~W. Kim, A.~Martin-Gomez, P.~Zhang, I.~Iordachita, and J.~U. Kang,
  ``Region targeted robotic needle guidance using a camera-integrated optical
  coherence tomography,'' in \emph{Optical Coherence Tomography}.\hskip 1em
  plus 0.5em minus 0.4em\relax Optica Publishing Group, 2022, pp. CM2E--6.

\bibitem{jagersand_vs}
M.~J. agersand and R.~C. Nelson, ``Adaptive differential visual feedback for
  uncalibrated hand-eye coordination and motor control,'' 1994.

\bibitem{Broydens_method}
C.~G. Broyden, ``A class of methods for solving nonlinear simultaneous
  equations,'' \emph{Mathematics of Computation}, vol.~19, pp. 577--593, 1965.

\bibitem{eye_surgery_imitation_learning}
J.~W. Kim, P.~Zhang, P.~L. Gehlbach, I.~I. Iordachita, and M.~Kobilarov,
  ``Towards autonomous eye surgery by combining deep imitation learning with
  optimal control,'' \emph{Proceedings of machine learning research}, vol. 155,
  pp. 2347--2358, 2020.

\bibitem{ronneberger2015u}
O.~Ronneberger, P.~Fischer, and T.~Brox, ``U-net: Convolutional networks for
  biomedical image segmentation,'' in \emph{Medical Image Computing and
  Computer-Assisted Intervention}.\hskip 1em plus 0.5em minus 0.4em\relax
  Springer, 2015, pp. 234--241.

\bibitem{he2016deep}
K.~He, X.~Zhang, S.~Ren, and J.~Sun, ``Deep residual learning for image
  recognition,'' in \emph{Proceedings of the IEEE conference on computer vision
  and pattern recognition}, 2016, pp. 770--778.

\bibitem{Steadyhand2}
A.~{\"U}neri, M.~A. Balicki, J.~Handa, P.~Gehlbach, R.~H. Taylor, and
  I.~Iordachita, ``New steady-hand eye robot with micro-force sensing for
  vitreoretinal surgery,'' in \emph{Biomedical Robotics and Biomechatronics
  (BioRob), 2010 3rd IEEE RAS and EMBS International Conference on}.\hskip 1em
  plus 0.5em minus 0.4em\relax IEEE, 2010, pp. 814--819.

\end{thebibliography}

\newpage

\vfill

\end{document}